\documentclass[manuscript,acmlarge]{acmart}
\usepackage{multirow}
\usepackage{tabularx}
\usepackage{caption}
\usepackage{subcaption}
\usepackage{algorithm}
\usepackage[noend]{algpseudocode}
\algrenewcommand\algorithmicrequire{\textbf{Input:}}
\algrenewcommand\algorithmicensure{\textbf{Output:}}

\def\algbackskip{\hskip-\ALG@thistlm}
\makeatother
\AtBeginDocument{%
  \providecommand\BibTeX{{%
    \normalfont B\kern-0.5em{\scshape i\kern-0.25em b}\kern-0.8em\TeX}}}

\setcopyright{acmlicensed}
\acmJournal{IMWUT}
\acmYear{2024} \acmVolume{8} \acmNumber{1} \acmArticle{10} \acmMonth{3}\acmDOI{10.1145/3643514}

\begin{document}

\title{MAPLE: Mobile App Prediction Leveraging Large Language Model Embeddings}


\author{Yonchanok Khaokaew}
\orcid{0000-0003-4297-6274}
\email{y.khaokaew@unsw.edu.au}
\affiliation{%
  \institution{School of Computer Science and Engineering, University of New South Wales }
    \city{Sydney}
  \state{NSW}
  \country{Australia}
  \postcode{2052}
}

\author{Hao Xue}
\orcid{0000-0003-1700-9215}
\email{hao.xue1@unsw.edu.au}
\affiliation{%
  \institution{School of Computer Science and Engineering, University of New South Wales }
  \city{Sydney}
  \state{NSW}
  \country{Australia}
  \postcode{2052}
}

\author{Flora D. Salim}
\orcid{0000-0002-1237-1664}
\email{flora.salim@unsw.edu.au}
\affiliation{%
  \institution{School of Computer Science and Engineering, University of New South Wales }
    \city{Sydney}
  \state{NSW}
  \country{Australia}
  \postcode{2052}
}

\renewcommand{\shortauthors}{Khaokaew Y., et al.}

\newcommand{\hlnewtwo}[1]{\textcolor{black}{#1}}
\newcommand{\hlnew}[1]{\textcolor{red}{#1}}
\newcommand{\hlcolor}[1]{\color{black}}

\begin{abstract}
In recent years, predicting mobile app usage has become increasingly important for areas like app recommendation, user behaviour analysis, and mobile resource management. Existing models, however, struggle with the heterogeneous nature of contextual data and the user cold start problem. This study introduces a novel prediction model, Mobile App Prediction Leveraging Large Language Model Embeddings (MAPLE), which employs Large Language Models (LLMs) and installed app similarity to overcome these challenges. MAPLE utilises the power of LLMs to process contextual data and discern intricate relationships within it effectively. Additionally, we explore the use of installed app similarity to address the cold start problem, facilitating the modelling of user preferences and habits, even for new users with limited historical data. In essence, our research presents MAPLE as a novel, potent, and practical approach to app usage prediction, making significant strides in resolving issues faced by existing models. MAPLE stands out as a comprehensive and effective solution, setting a new benchmark for more precise and personalised app usage predictions. In tests on two real-world datasets, MAPLE surpasses contemporary models in both standard and cold start scenarios. These outcomes validate MAPLE’s capacity for precise app usage predictions and its resilience against the cold start problem. This enhanced performance stems from the model’s proficiency in capturing complex temporal patterns and leveraging contextual information. As a result, MAPLE can potentially improve personalised mobile app usage predictions and user experiences markedly.
\end{abstract}




\received{15 May 2023}
\received[revised]{15 November 2023}
\received[accepted]{11 January 2024}

\maketitle

\section{INTRODUCTION}\label{sec:intro}

\hlcolor

As reliance on smartphones increases, predicting app usage has become crucial. Smartphones are utilised for various tasks, including communication, entertainment, and work. Accurate predictions of app usage can enhance user experience, conserve battery life, and improve network efficiency. Furthermore, app usage prediction can aid app developers and marketers in better understanding user behaviour, allowing them to tailor their offerings more effectively to their target audience.

Previous research has underscored the importance of contextual information in app usage predictions, encompassing aspects like time, location, and user history. Nevertheless, integrating this contextual data presents challenges. Embedding such information requires extensive training, which can be time-consuming and resource-intensive. Additionally, incorporating this data into predictive models can be complicated, especially when dealing with the cold start problem.

Recent advancements in natural language processing have led to the development of Large Language Models (LLMs) based on the Transformer architecture. These models, such as GPT-3 \cite{brown2020languagegpt3}, demonstrate remarkable proficiency in various natural language tasks, including language generation, translation, and sentiment analysis. Their flexibility and adaptability make LLMs well-suited for diverse domains, including app usage prediction. The Transformer architecture \cite{vaswani2017attention}, foundational to many state-of-the-art LLMs, was initially designed for tasks involving sequences in natural language processing, making it particularly apt for app usage prediction.

LLMs are especially effective in processing text data, positioning them as prime candidates for predicting app usage. Contextual information, such as location, can be textually represented (e.g., "at a coffee shop") or behaviours (e.g., "frequently uses social media apps in the evening"). Utilizing LLMs and the Transformer architecture enhances the precision of app usage predictions. These models leverage features like self-attention and multi-head attention, effectively capturing app usage sequences in a manner similar to natural language processing. Beyond improving embedding training, LLMs address the cold start problem associated with new contexts or users. Traditional models frequently encounter difficulties when dealing with unfamiliar contexts. However, Large Language Models (LLMs), utilizing their vast pre-trained language abilities, are adept at generating pertinent text descriptions, thereby enhancing prediction accuracy. To address the issue of predicting for new users who lack historical data, our approach utilises patterns observed in users with similar app usage. This strategy proficiently handles cold start scenarios, effectively circumventing the limitations typically associated with conventional embedding training methods.

\color{black}

Leveraging insights from our experiments, we propose a novel app usage prediction model that capitalizes on the strengths of LLMs and contextual data to tackle the user cold start issue for new users. This model employs LLMs to translate contextual information, such as time and location, into text format. It also integrates behaviour data from users with a history of app usage, enhancing prediction accuracy. A key advantage of using LLMs is the elimination of the need for initial training of the embedding layer. We utilise the pre-trained embeddings from LLMs, significantly reducing computational demands and training time.

Our model, named \hlnewtwo{Mobile App Prediction Leveraging Large Language Model Embeddings (MAPLE)}, primarily harnesses LLMs to process contextual data, yielding a more refined understanding of user behaviour. Additionally, it incorporates behaviour patterns from users with similar app usage to address the cold start problem in new users, enhancing both accuracy and generalizability. By identifying and analyzing users with analogous app usage patterns, MAPLE can discern underlying behaviour trends, leading to more personalised and precise app usage predictions. In essence, our model's contributions are threefold:

\begin{itemize}
    \item The proposed model leverages LLMs to process contextual information, which provides a more accurate and nuanced understanding of user behaviour.
\item The model addresses the user cold start problem for new users by incorporating the behaviour data of similar users with a history of app usage, improving the model's accuracy and generalizability.
\item By utilizing pre-trained embeddings of LLMs, the proposed model significantly reduces computational requirements and training time, making it more efficient and scalable.
\end{itemize}

In the following sections, we will delve deeper into the dataset and problem statement, present our language model-based methodology, evaluate the performance of our model using real-world datasets, and discuss the implications of our findings for the future of app development and user experience optimization.

\section{MOTIVATION AND PROBLEM DEFINITION}

\subsection{Motivation}
The motivation for this research is rooted in addressing the significant challenges associated with contextual data heterogeneity and the cold start problem in app usage prediction. The diverse and complex nature of contextual data hinders the development of accurate and robust models for predicting user behaviour. Furthermore, the cold start problem exacerbates these challenges, as providing accurate predictions for new users with limited historical data remains daunting.

\hlnewtwo{
Recognizing the capabilities of Large Language Models (LLMs) and their foundational Transformer architecture, our research seeks to utilise their strengths in managing the complexity of contextual data. LLMs have shown exceptional performance in diverse natural language processing tasks, and recent studies indicate their effectiveness in time series forecasting as well \cite{xue2022leveragingmobicast, xue2022translating}. Given their proficiency in handling and generating sequential data, crucial for understanding app usage behaviour, we are motivated to investigate their use in app usage prediction. This domain aligns with time series forecasting in terms of identifying sequential patterns. By deploying LLMs, we aim to leverage their advanced processing power to unravel complex relationships within contextual data, thereby enhancing the accuracy of app usage predictions.}

Furthermore, we aim to explore the possibility of using the similarity of installed apps as a marker of comparable app usage behaviours. This approach could effectively address the cold start problem and facilitate the modelling of user preferences and habits, even for new users with limited historical data. By integrating the advantages of Large Language Models (LLMs) with insights gained from installed app similarity, we seek to provide a thorough and effective method for predicting app usage. Our research stands to make a substantial contribution to the field, presenting a novel, robust, and practical approach that tackles key challenges faced by current models in app usage prediction.

\hlcolor

\subsubsection{Contextual Information Used in Modelling Mobile User Behaviour}
Contextual information is pivotal in modelling mobile user behaviour, and a range of contextual factors have been employed in previous research to predict user behaviour effectively. The types of contextual information frequently utilised in modelling mobile user behaviour include the following.
\begin{itemize}
    \item Location: User location significantly influences behaviour predictions. Studies show that a user's current location, their location during app usage, and frequent visit spots offer behaviour insights. \cite{xia2020deepapp,suleiman2021deeppatterns,Cap,SA-GCN}.
    \item Previous Applications: Prior studies have shown that previously used applications can be indicative of the user's future application use \cite{xia2020deepapp,Cap,appusage2vec,kang2022app,suleiman2021deeppatterns,khaokaew2021cosem}. The extraction of previous application usage as a feature in prior works has demonstrated its usefulness in predicting mobile user behaviour.
    \item App Categories: The user's applications category can provide insight into their interests and preferences. By analyzing the types of applications a user frequently uses, such as social media, productivity, or entertainment, we can better understand their interests and predict their future behaviour \cite{Cap,SA-GCN}.
    \item Temporal Information: The day of the week, hour of the day, and day of the month that a user used an application can provide insight into their preferences and usage patterns. Prior works have shown that the time of day when a user most frequently uses a specific application and their typical usage pattern throughout the day can provide further information \cite{aliannejadi2021context,khaokaew2021cosem,xia2020deepapp,suleiman2021deeppatterns,appusage2vec}.
    \item Anonymized ID: A unique identifier helps track individual usage patterns. Previous research suggests age and gender also illuminate user behaviour \cite{appusage2vec,xia2020deepapp,suleiman2021deeppatterns}. However, due to privacy and cold start problems, our model excludes anonymous IDs, as their absence in cold start scenarios would limit the model's relevance.
\end{itemize}

\begin{figure}[!h]

  \includegraphics[width=0.75\linewidth]{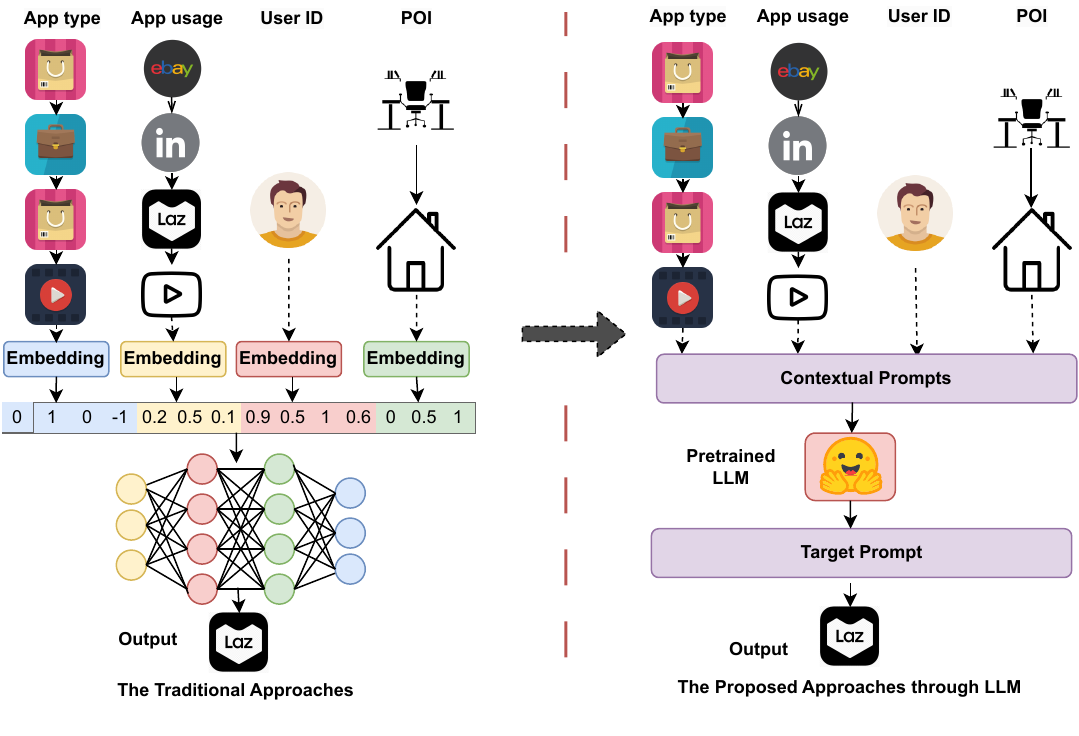}
  \caption{Conceptual illustrations of traditional approaches for using contextual information in app usage prediction problems compared to the proposed model based on LLMs}
  \label{fig:old embedding}
\end{figure}

\color{black}
\hlnewtwo{In traditional methods of modelling mobile user behaviour, embedded layers represent contextual information \ref{fig:old embedding}.} Yet, these approaches face challenges with diverse and evolving contextual data. This is due to the static nature of embedded layers, which struggle to adapt to new or changing contexts. Moreover, for each new user or distinct context type, these layers require retraining, a process that can be both time-consuming and demanding in terms of resources.

Applying a Large Language Model (LLM) presents a solution that addresses these limitations, offering a more expansive and adaptable approach. Being a pre-trained model, the LLM can be fine-tuned for specific applications, like predicting mobile user behaviour. It effectively processes various contextual information and seamlessly adjusts to new and evolving contexts. In contrast, traditional methods of modelling mobile user behaviour struggle to manage such a broad spectrum of contextual data and to make predictions based on it. Employing an LLM provides a way to surmount these challenges, leading to a more comprehensive and efficacious method for modelling mobile user behaviour.

The table \ref{tab:prompt contexct} encapsulates the contextual information typically used in modelling mobile user behaviour, alongside potential prompts for extracting pertinent data from each context type. These prompts, derived from the accessible contextual data, offer a targeted means for the LLM to discern relevant information and forecast mobile user behaviour. By formulating these prompts based on the available contextual information, the LLM is equipped to parse semantic details, thereby enhancing its accuracy in predicting mobile user behaviour.

\begin{table}[!h]
  \caption{The possible prompts that can be applied to each type of context information}
  \label{tab:prompt contexct}
  \small
  \begin{tabular}{c|p{9 cm}}
    \toprule
    Contextual Information& Possible Prompts\\
     \midrule
    Location&"The user is currently at [LOCATION]." \newline "The user visited [LOCATION] when using a specific app." \newline "The user frequently visits [LOCATION]." \newline "The user is close to [POINT OF INTEREST]."\\
    \midrule
    Previous Applications&"The user has recently used [APP]." \newline "The user frequently uses [APP]." \newline "The app [APP] is used recently."\\
    \midrule
    App Categories&"The user frequently uses apps in the [CATEGORY] category." \newline "The app [CATEGORY] is used prior to the prediction."\\ \midrule

    Temporal Information&	"It is currently [DAY OF WEEK]." \newline "It is [HOUR OF DAY]." \newline "It is [DAY OF MONTH]." \newline "The user frequently uses [APP] at [TIME OF DAY]." \newline "It is [TIME OF DAY ] on [WEEK DAY]." \\ \midrule
Anonymised ID	& "The user [Anonymised ID]'s age is [AGE]." \newline "The user [Anonymised ID]'s gender is [GENDER]."\\ \midrule

\end{tabular}
\end{table}

In our proposed approach, LLMs will be utilised in conjunction with contextual information and prompts derived from this data. These prompts enable the LLM to discern relevant details from the given contextual information, leading to a more thorough and effective method for predicting mobile user behaviour. The synergistic use of LLMs with prompts based on contextual information represents an innovative strategy for modelling mobile user behaviour. This technique has the potential to yield more precise predictions and deeper insights into user behaviour. Our research is dedicated to examining this approach and its capacity to offer a more comprehensive and efficacious solution, addressing the limitations of traditional methods in modelling mobile user behaviour.

\subsubsection{Overcoming the User Cold Start Problem with Installed Application Analysis}\label{sec:coldstart}

The user cold start problem refers to the difficulty in predicting a new user's behaviour due to the lack of available data. In the context of mobile user behaviour, this can make it challenging to accurately predict the applications that a new user is likely to use, especially when the user's identifier, which is commonly used as contextual information in modelling mobile user behaviour \cite{appusage2vec,xia2020deepapp,suleiman2021deeppatterns}, is not available for the new user.

\hlcolor

To address this issue, we aim to investigate the relationship between the set of installed applications and mobile user behaviour, specifically app usage, in this section. By analyzing the installed applications, we hope to gain insight into how they can indicate a user's interests and habits and how this information can be used to predict their future app usage. This analysis is crucial in overcoming the user cold start problem, as it provides a way to make initial predictions about a new user's behaviour, even when the user's identifier is unavailable. These predictions can then be refined and improved over time as more data about the user becomes available.

 First, we'll employ Jaccard Similarity to identify users with analogous app sets, aiming to understand how app collections reflect interests and predict future usage. Using Principal Component Analysis (PCA) and t-Distributed Neighbor Embedding (t-SNE) \cite{li2023hierarchical,van2008visualizing}, we'll condense the one-hot encoded vectors of selected users' app categories for easier two-dimensional visualization of app usage trends.

\begin{figure}[!h]
     \begin{subfigure}[b]{0.44\textwidth}
         \centering
           \includegraphics[width=\textwidth]{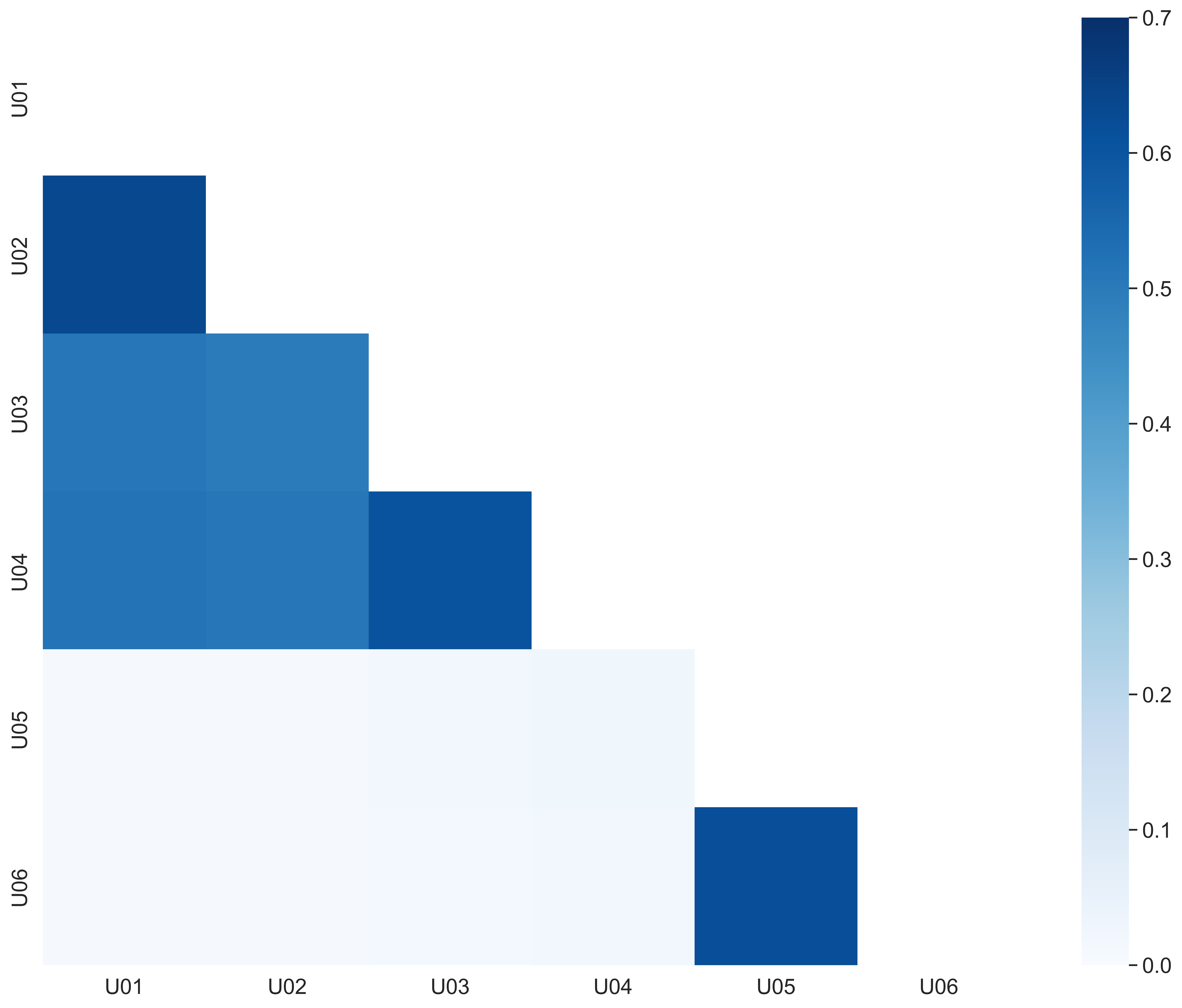}
         \caption{Jaccard Similarity Heatmap of Installed Apps Among Top Three Similar Users in Tsinghua App Usage Dataset}
         \label{fig:jaccard}
     \end{subfigure}
          \begin{subfigure}[b]{0.44\textwidth}
         \centering
           \includegraphics[width=\textwidth]{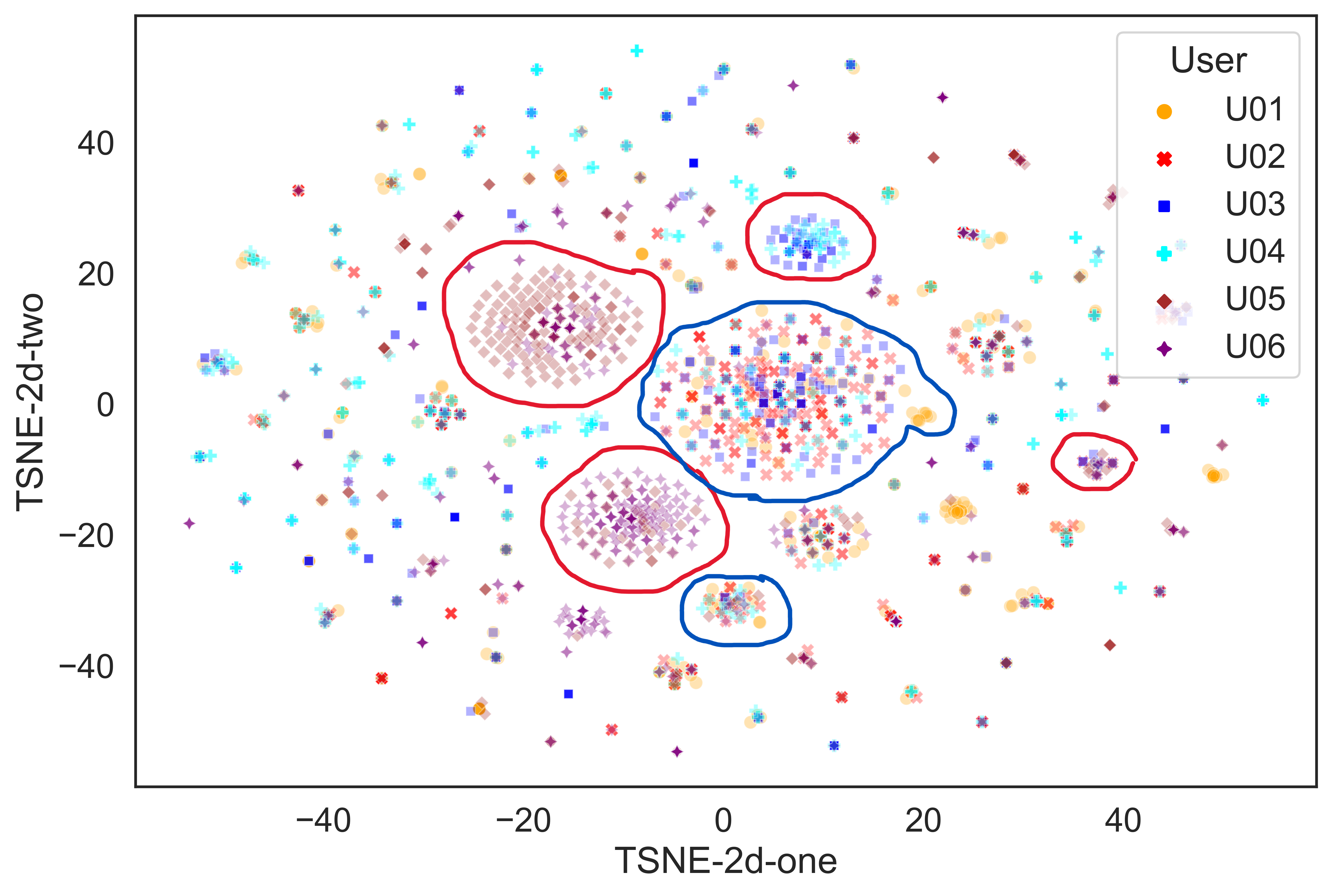}
         \caption{t-SNE Reduced 2D Visualization of App Usage Behaviour Among Users with Similar Installed App Sets}
         \label{fig:gu-line}
     \end{subfigure}

  \caption{Exploring App Usage Behaviour Similarity Among Users with Similar Installed App Sets}
  \label{fig:Ana2.2}
\end{figure}

\hlnewtwo{Figure \ref{fig:Ana2.2} depicts our findings. The heatmap (Fig \ref{fig:jaccard}) shows Jaccard Similarity scores among user pairs, ranging from 0 (dissimilar) to 1 (identical). Three user pairs—U1 \& U2, U3 \& U4, and U5 \& U6—demonstrate high similarity over 0.6, suggesting comparable app interests and behaviours. Notably, U3 \& U4 share moderate similarity (around 0.4) with U1 \& U2, hinting at parallel behaviours. This analysis precedes a detailed visualization of these users' app usage patterns.}

The scatter chart \ref{fig:gu-line} analyzes users' app usage with similar installed apps, identified through Jaccard Similarity. Displayed in a two-dimensional space with axes representing the principal components from t-SNE, each point signifies a user's app usage, colour-coded for distinction. Notably, clusters, especially the red circles for users U5, U6, U3, and U4, signify shared app behaviour due to similar app sets.

Furthermore, a blue circle in the chart highlights four users, U1, U2, U3, and U4, who have similar apps and behaviours, emphasizing the connection between the installed apps and user inclinations. This observation is pivotal for forecasting future app usage. The chart expands upon the heatmap's analysis by illustrating the relationships between apps and user similarities. These insights will guide the development of our predictive model for future app usage, concentrating on installed apps to address the user cold start problem. Next, we will articulate the problem formulation for our prediction model, integrating installed apps with contextual data as a foundation for its development.

\color{black}

\subsection{Problem Definition}

\hlcolor

In mobile user behaviour modelling, predicting future app usage is essential. A user's app usage is influenced by various contextual factors such as app usage history, time, and location. To predict a user's next app, it's vital to integrate these factors into the model.

We examine contextual information types $C = {A,J, T, L}$, where $A$ denotes app usage history, $J$ app type history, $T$ time, and $L$ location. These types are converted into sentences $S_c = {S_A, S_J, S_T, S_L}$ to represent a user's context fully. The goal is to predict the next app $\mathcal{P}$ a user will use, through a machine learning model $z$, formulated as:

\begin{equation}
z(S_c) \rightarrow \mathcal{P}
\end{equation}

Here, $z(\cdot)$ is the function transforming contextual sentences $S_c$ into the next app prediction $\mathcal{P}$. We employ a large language model (LLM) to address the shortcomings of traditional embedding models and capture the user's context accurately. We'll train and test $z(\cdot)$ on real-world data to evaluate its predictive accuracy with contextual information. Our ultimate aim is to create a model that leverages contextual data and installed apps to accurately forecast a user's app usage, offering insights into mobile user behaviour modelling and enhancing the field's progress.

\color{black}

\section{Dataset}
This section is dedicated to a thorough analysis of the datasets used in our study. We have employed two publicly accessible app usage datasets, chosen for their relevance to our research question. Our analysis includes both statistical and characteristic examinations to gain a comprehensive understanding of the data. The statistical analysis involves calculating essential descriptive statistics providing an overview of data distribution. In contrast, the characteristic analysis offers a deeper exploration of the data's features, trends, and patterns. This includes evaluating the overall characteristics and noting differences between the two datasets.

\subsection{Tsinghua App Usage Dataset}
The first dataset, comprising app usage records, is sourced via Deep Packet Inspection (DPI) devices, a leading mobile network operator in China. This dataset records mobile users’ app usage, capturing both time and location details as users connect to the cellular network. The locations in the dataset reflect the accuracy of the cell tower's position. Each record in the dataset contains an anonymized user ID, timestamps of HTTP requests or responses, packet lengths, visited domains, and the app's ID. The data was collected in Shanghai, a major metropolitan area in China. A summary of the dataset's statistics for Tsinghua App Usage is presented in Table \ref{tab:statistic datasets}.

\begin{table}[!h]
  \caption{Dataset Statistics for Tsinghua App Usage and LSApp}
  \label{tab:statistic datasets}
  \begin{tabular}{l|c|c}
    \toprule
    Metric& Tsinghua App Usage&LSApp\\
    \midrule

    \# users& 870 & 292\\
    \# unique app& 2,000 & 87\\
    \# sessions&102,422& 76,247\\
        \# locations& 6,560& N/A\\
    \# app usage records    &  ~2.4 M  &           ~600 K\\
    
     Mean unique apps in each session & 3.50 ± 7.75  &   2.18 ± 1.46  \\
     Mean used apps per location& 24.96 &N/A \\
  \bottomrule
\end{tabular}
\end{table}

\subsection{Large Dataset of Sequential Mobile App Usage (LSApp)}
LSApp \cite{aliannejadi2021context}, serving as a supplementary dataset to ISTAS \cite{khaokaew2021cosem,Aliannejadi.CIKM201810.1145/3269206.3271679}, similarly compiles sequential app usage events from users. In this study, 292 participants were enrolled and instructed to install the uSearch application on their smartphones, operating it for a minimum of 24 hours. Participants were also asked to log their mobile searches in uSearch as they occurred promptly. Additionally, app usage statistics were gathered with the consent of the participants. The dataset's statistics for LSApp are detailed in Table. \ref{tab:statistic datasets}.

\subsection{Dataset Characteristics and Analysis}

\hlcolor

Analyzing the characteristics of the datasets used in a study is crucial in ensuring the validity and generalizability of the results. In this section, we analyse the characteristics of the Tsinghua App Usage and LSApp datasets mentioned in the previous section.

 \begin{table*}[!h]
     \centering
      \caption{Dataset and Available contextual information }
     \begin{tabular}{c|c|c|c|c}
     Dataset & Historical app usage & Point of interest & Temporal info. & App type info.\\
     \hline
        Tsinghua App Usage dataset  & $\times$& $\times$& $\times$ &$\times$ \\
        LSApp dataset  & $\times$ & & $\times$ &$\times$ \\
     \end{tabular}
    
     \label{tab:dataset-context}
 \end{table*}

Table \ref{tab:dataset-context} outlines the contextual data in each dataset. Both the Tsinghua App Usage and LSApp datasets include historical app usage, application types, temporal data, and user IDs. However, the Tsinghua App Usage dataset offers an extra layer of context with points of interest, potentially enhancing prediction model accuracy.

Our proposed model employs a Large Language Model (LLM) adaptable to the data available, whether it's historical app usage, temporal information, or additional data like the points of interest in the Tsinghua App Usage dataset. This adaptability makes the model versatile across various datasets.

\begin{figure}[!h]
\begin{subfigure}[b]{0.65\textwidth}
\includegraphics[width=\linewidth]{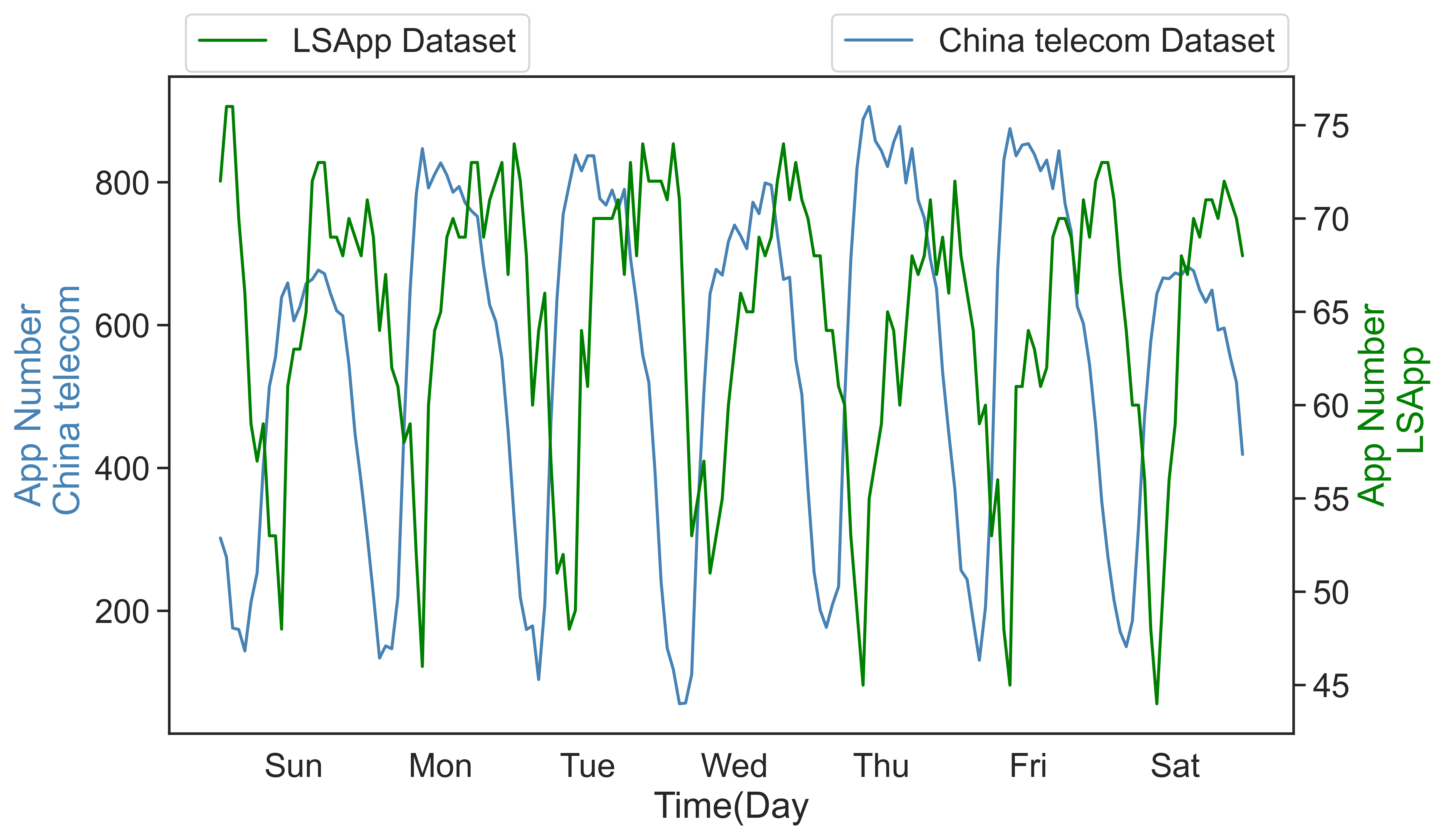}
\caption{Temporal distribution of total mobile App usage number (1 hours resolution)}
  \label{fig:Temporal distribution}
\end{subfigure}
     \begin{subfigure}[b]{0.48\textwidth}
         \centering
           \includegraphics[width=\textwidth]{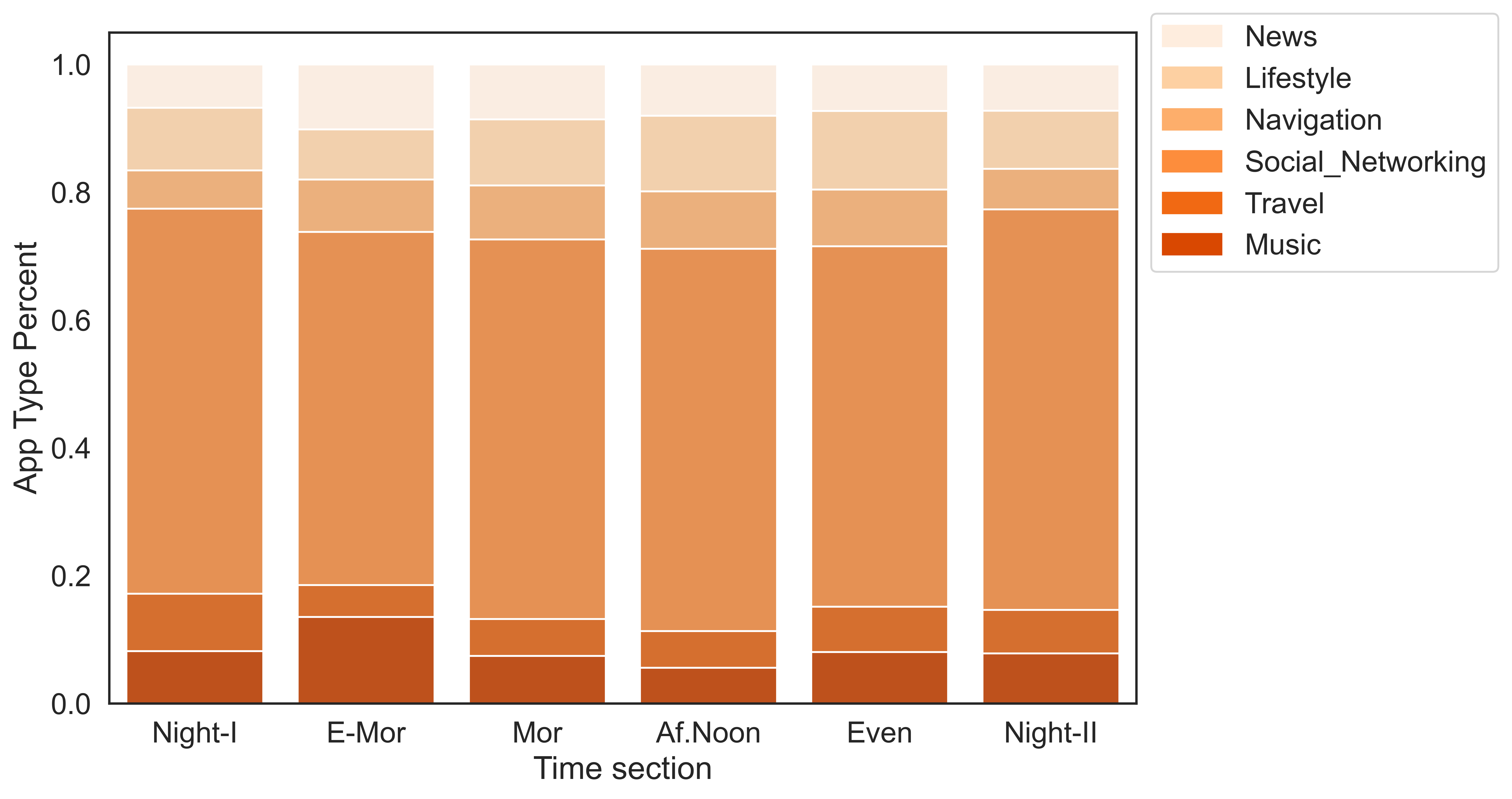}
         \caption{App type usage distribution of Tsinghua App Usage Dataset}
         \label{fig:ac_ch ids}
     \end{subfigure}
          \begin{subfigure}[b]{0.48\textwidth}
         \centering
           \includegraphics[width=\textwidth]{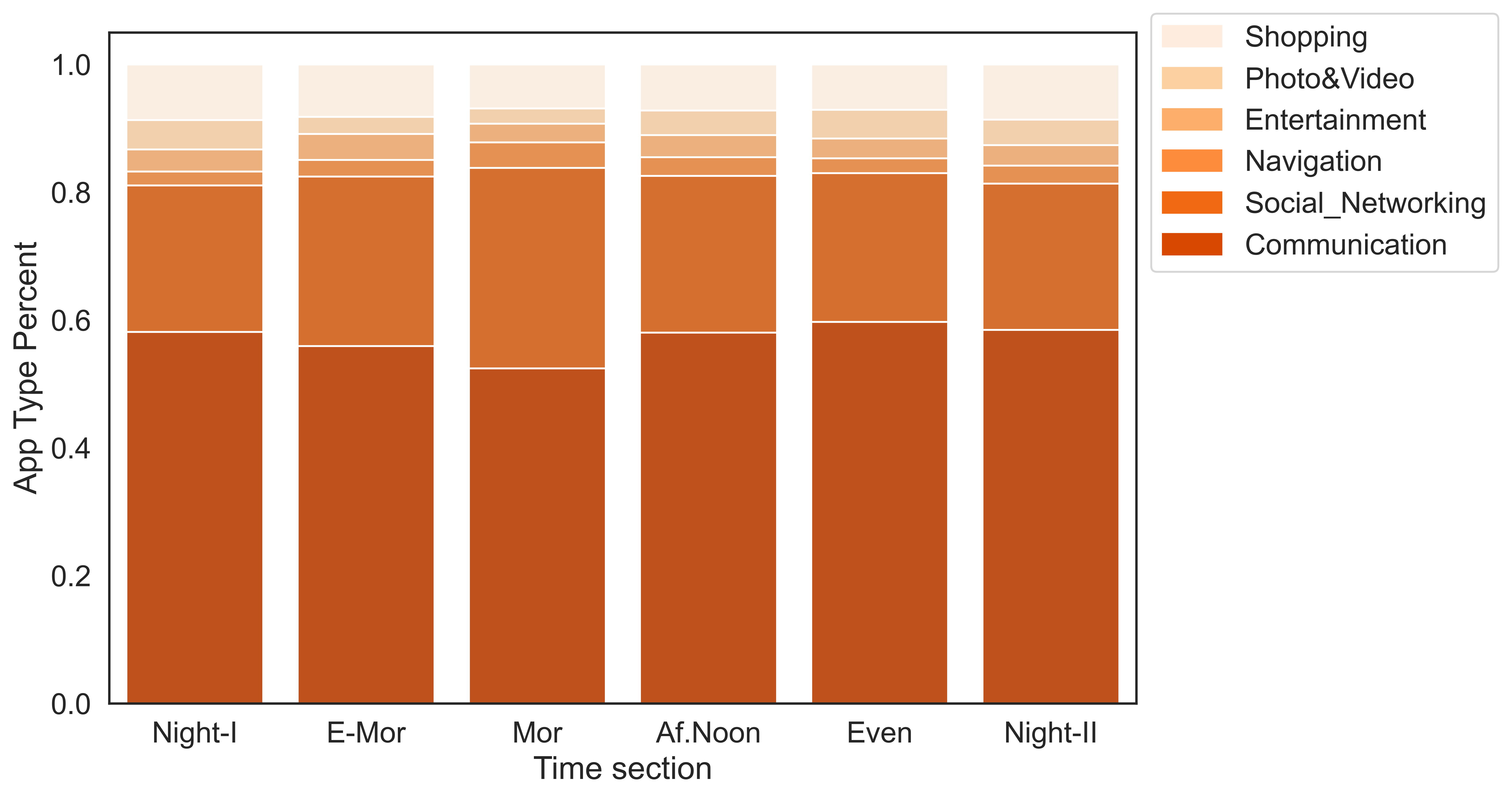}
         \caption{App type usage distribution of LSApp Dataset}
         \label{fig:ac_cls ids}
     \end{subfigure}

  \caption{The usage of Apps at different time segments from different datasets.}
  \label{fig:appstacked}
\end{figure}

We analyzed app usage over various time frames, as shown in Figure \ref{fig:Temporal distribution}. The x-axis represents time in 1-hour intervals from Sunday to Saturday, and the y-axis indicates the number of app usages in each interval. This graph compares app usage trends over a week, employing dual y-axes to contrast the Tsinghua App Usage and LSApp datasets. Both datasets exhibit similar patterns, with increased activity in the afternoons and evenings and reduced usage in early mornings and late nights. A noticeable difference is observed in the peak times: the Tsinghua App Usage dataset shows an earlier peak than the LSApp dataset. Utilizing an LLM model can adeptly address these temporal variations. Owing to its extensive pre-training, LLMs are skilled at identifying patterns across diverse datasets. By processing context in sentence form, LLMs adeptly discern each dataset's unique app usage trends, thereby facilitating accurate predictions.

We also examined app type usage across various time segments (Night I, E-morning, Morning, Afternoon, Evening, Night II) from two datasets, as depicted in Fig. \ref{fig:appstacked}. The Tsinghua App Usage dataset predominantly uses social network apps, while the LSApp dataset primarily uses communication apps, with social networks as a close second. Navigation apps rank in the top 6 in both, peaking in the mornings and afternoons. Coupled with insights from Figure \ref{fig:Temporal distribution}, its explicit app usage varies by time. Thus, integrating app history and temporal details can refine prediction accuracy. Recognizing this time-dependent app usage, it's vital to devise a model capturing these nuances. Given its adeptness at handling intricate data dynamics, we believe a large language model can address this. By fusing historical, temporal, and location data, our model aims to decipher context-driven app usage patterns better. The following sections will describe our proposed model and evaluate its performance using real-world datasets.

\color{black}

\section{ PROPOSED MODEL}

\subsection{Model overview}
\hlcolor

\begin{figure}[h!]

  \includegraphics[width=0.85\linewidth]{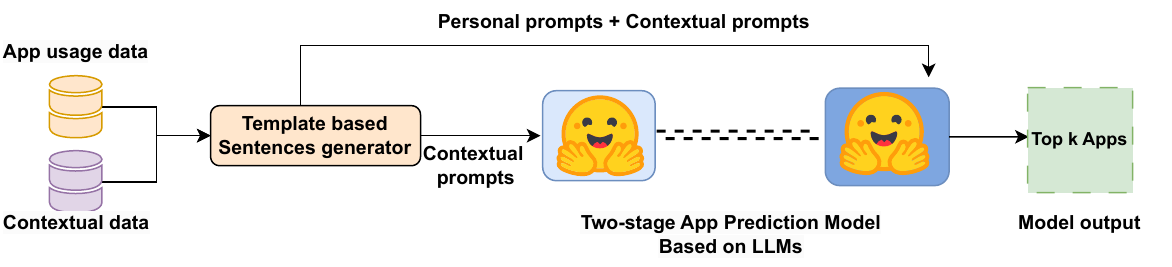}
  \caption{Overall flowchart of the proposed model}
  \label{fig:OVHAM}
\end{figure}

This section introduces a proposed methodology to tackle the challenges outlined in Section \ref{sec:intro}. Our proposed model, named Mobile App Prediction Leveraging Large Language model Embeddings (MAPLE), depicted in Figure.\ref{fig:OVHAM}, is crafted to scrutinize mobile user behaviour and predict app usage accurately, utilizing the capabilities of a large language model. MAPLE leverages installed app data to discern new users' interests, habits, and preferences, thus addressing the user cold start issue. The model is structured around two main components.

The initial component, the Template-based Contextual Sentence Generation Module, processes app and contextual information to generate sentences that encapsulate the user's present mobile usage context. These context-rich sentences are formulated using templates to integrate key information like time of day, location, and app type. Furthermore, we utilise the capabilities of Large Language Models in our distinctive Two-stage App Prediction Model. This approach unfolds sequentially: Initially, it focuses on predicting the broad category of the app. Subsequently, our model refines its prediction, narrowing down to the specific app a user will most likely use next.

In essence, the proposed model harnesses the prowess of language models to adeptly handle contextual data heterogeneity and address the user cold start problem. Subsequent subsections will delve into more intricate details of each model component, showcasing our expert approach.

\color{black}

\begin{table*}[h!]

  \caption{The template for converting available contextual information to the written language used by the proposed prediction modules.}

  \label{tab:template_combined}
  \begin{tabular}{p{1cm}|p{2.3cm}|p{5.5cm}|p{6.7cm}}
    \toprule
    &  Description & Template & Example \\
    \toprule
    \multirow{7}{=}{Input}&  
    Historical app \newline category usage& The apps \{$c_1,c_2,....,c_{t_{obs-1}}$\} are used prior to the prediction. &  The apps Photo/Video, Communication, and Utilities are used prior to the prediction.  \\\cmidrule{2-4}
    & Prediction Time& On \{$a_{t_{obs}}$\}& On Tuesday 02 PM\\\cmidrule{2-4}
    & Point of interest&The user is close to \{$l_{t_{1}}$\}, \{$l_{t_{2}}$\} and $\{l_{t_{obs-1}}$\}. &The user is close to service, shopping and restaurants.\\
    \\\cmidrule{2-4}
    &Historical app \newline usage & The apps \{$a_1,a_2,....,a_{t_{obs-1}}$\} are used prior to the prediction. & The apps 1, 4, and 9 are used prior to the prediction.  \\\cmidrule{2-4}
    &Installed apps & $c_1 : a_{1_{c_1}},a_{2_{c_1}}$\newline$c_2:a_{1_{c_2}},a_{2_{c_2}}$ ..... ..... \newline & travel apps : 1,4,12  \newline
    utility apps : 2,7,16  \\
    
    \midrule
    \multirow{2}{=}{Output} & $1^{st}$ stage result & Based on the global information, the next app will be a \{$c_{1t_{obs}}$\} app (\{$p_{1t_{obs}}$\}\%), \{$c_{2t_{obs}}$\} app (\{$p_{2t_{obs}}$\}\%) or \{$c_{3t_{obs}}$\} app (\{$p_{1t_{obs}}$\}\%)  & Based on the global information, the next app will be a communication app (70\%), social app (20\%) or travel app (10\%) \\\cmidrule{2-4}
    & $2^{nd}$ stage result &  This user will use App {$a_{t_{obs}}$}.  & This user will use App 4. \\

  \bottomrule
\end{tabular}

\end{table*}

\subsection{Template-based contextual sentence generation}

\hlcolor

The Template-based Contextual Sentence Generation Module stands as a pivotal element in our LLM-based proposed method. Its primary function is to transform available context features into informative and coherent sentences, which serve as inputs for the downstream prediction modules. This module is essential for crafting contextual sentences that encapsulate the user’s current mobile usage context, thus supplying vital contextual data to the subsequent components of our model. Drawing inspiration from the work of \citet{xue2022translating}, we have devised a template-based strategy to create these contextual sentences. Our method encompasses two distinct template types, each tailored for extracting specific information types: one for the App Type Prediction Training Stage (ATP Training Stage) and another for the Next App Prediction Training Stage (NAP Training Stage). Utilizing these templates, our model can produce coherent, context-relevant sentences that not only accurately foresee the next app type but also offer insights into the user’s mobile usage patterns. Details of the template for converting contextual information into textual format are illustrated in Table \ref{tab:template_combined}.

To construct contextual sentences for the ATP and NAP Training Stages, we utilise predefined templates tailored to our dataset’s contextual data and adaptable for specific use cases. These stages incorporate historical app usage categories, the time of prediction, and, when pertinent, the user's point of interest. The historical category denotes previously engaged apps, while the prediction time signifies the moment of prediction. The point of interest, linked to the user’s location, is integrated as needed. These templates are instrumental in generating sentences that accurately reflect the user's current context, thereby facilitating precise app category predictions. In the NAP Training Stage, features such as historical app usage, installed apps (as discussed in Section \ref{sec:coldstart}), and context from the ATP stage are employed in sentence generation. These elements are crucial for understanding user behaviour and preferences in anticipating the next app the user will use. The forthcoming sections will expound upon the ATP and NAP Training Stages.

\subsection{Two-Stage LLM Training Module}

\begin{figure}[h!]

  \includegraphics[width=0.75\linewidth]{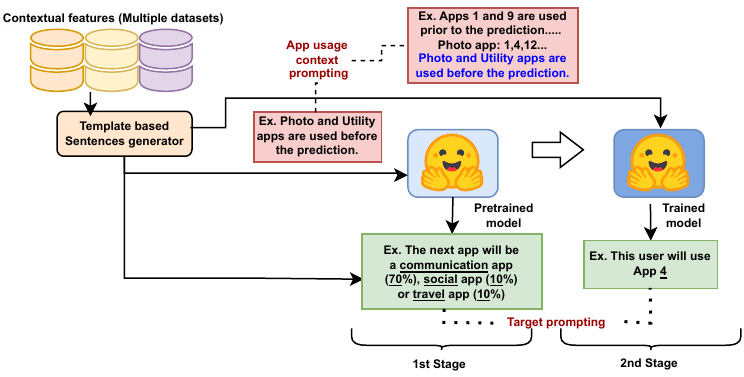}
  \caption{Overall flowchart of the two-stage App Prediction Model}
  \label{fig:OVmixed}
\end{figure}

\hlnewtwo{
Our MAPLE model's core is an innovative two-stage training process utilizing Large Language Models (LLMs), which goes beyond traditional seq2seq applications. This bespoke method, comprising the App Type Prediction Training Stage followed by the Next App Prediction Training Stage, is tailored to enhance the predictive precision of the next app a user will likely engage with.}

\subsubsection{App Type Prediction Training Stage (ATP)}

The first stage, the ATP Training Stage,  focuses on predicting the type of the next app based on historical app categories and contextual features. Harnessing the capabilities of Large Language Models (LLMs), this stage incorporates previous app categories, time, and location data to make predictions. Designed as a seq2seq model, a type of summarization task, this stage aims to generate target sentences based on the given contextual sentences  (as shown in Figure \ref{fig:OVmixed}). The mathematical representation of the seq2seq model is denoted as $S_{at} = Seq2Seq(S_{c_{at}}; \theta_)$, wherein $S_{c_{at}}$ denotes the contextual sentences for app type prediction, and $S_{at}$ represents the target sentence.

Firstly, we transform contexts into sentences using predefined templates. This stage amplifies the model's ability to predict subsequent app types by drawing insights from historical app usage data across all available datasets. We generate target sentences ($S_{at}$)  by calculating the probability of the next app category based on the given historical app sequence. This probability is derived from analyzing diverse app-type sequences that precede the target app type, as outlined in Algorithm \ref{alg:cap}. The resulting target sentences accurately capture users' app usage patterns, thereby enhancing the model's predictive capabilities.

\begin{algorithm}[!h]

\caption{Generate the next application type prompting}\label{alg:cap}

\begin{algorithmic}[1]

\Require Set of app type sequences $G$, number of datasets $D$, $k$ top app type 
\Ensure Next app categories prompting $S_{at}$

\State Initialize a dictionary final\_dict to store the probabilities for each app type sequence;
\For{each app type sequence $g$ $(Ac_1, Ac_2, ..., Ac_{n-1})$ in $G$}

\State Init. dict. \textit{count\_dict} to store the counts of app categories;
\For{each dataset $d$ in $D$}
\For{each user $u$ in $d$}
\State Iterate through the app type sequences of $u$;
\If{the sequence $g$ is found}
\State Incr. the count of the next app type $(Ac_n)$ in \textit{count\_dict};
\EndIf

\EndFor
\EndFor

\State  Calculate the sum of all counts in \textit{count\_dict} as \textit{total\_count};
\State  Init. a dict. \textit{prob\_dict} to store the prob. of app cats.;
\For{each app category $c$ in \textit{count\_dict}}
\State Calc. prob. $P(c | A_{1,..,n}) = count\_dict[c]/total\_count$;
\State Store the probability in \textit{prob\_dict}[$c$];
\EndFor
\State Sel. top $k$ app types with max. probs in \textit{p\_dict} \& store in \textit{top\_k\_dict};
\State Gen. next app type prompt from top $k$ types using template in Table \ref{tab:template_combined};

\State Add this generated sentence to \textit{final\_dict} with the key $g$;
\EndFor
\Return \textit{final\_dict};

\end{algorithmic}
\end{algorithm}

In sum, the ATP Training Stage enhances the model's proficiency in predicting app types by leveraging historical app categories and contextual features. It emphasizes learning from various contextual feature sentences and anticipates further improvement in the subsequent training stage.

\subsubsection{Next App Prediction Training Stage (NTP)}

The NTP Training Stage constitutes a pivotal advancement in our proposed model, building upon the insights gleaned from the App Type Prediction Training Stage. This stage aims to enhance the model's predictive capabilities further, enabling it to forecast the precise next app while considering broader contextual features. These encompass historical app types, individual app usage patterns, and the user's installed app repertoire.

We initiate this stage by leveraging the model trained during the ATP Training Stage. The seq2seq task (summarization) remains integral for predictions, and the model is enriched with a broader array of contextual features. These features encompass the contextual sentences from the preceding stage and additional personal context elements, such as individual app usage history and the user's installed app set. This enhancement empowers the model to understand and adapt to users' distinct habits and preferences. The seq2seq model for this stage is represented as: $S_{na} = Seq2Seq(S_{c_{na}}; \theta)$. Here, $S_{c_{na}}$ represents the amalgamation of contextual sentences used to predict the next app. It incorporates sentences from the previous stage and supplementary personal context features, including the installed app set and historical app ID sequence. ${S_{na}}$ denotes the target sentence ($2^{nd}$ stage). This enriched context significantly contributes to predicting the next app, thus heightening predictive accuracy.

The NTP Training Stage, illustrated in Figure \ref{fig:OVmixed}, merges installed app data with contextual information to formulate the next app prompts. Leveraging LLMs, this stage models complex relationships between installed apps and contextual data, enabling highly accurate predictions of subsequent app usage. The prompts propose the most suitable next app based on user history and context.

This stage essentially amalgamates personal app history, installed app set, historical app type usage, and contextual information to yield exact predictions of the next app. This stage becomes pivotal in mobile app development by utilising LLM-based modelling and seq2seq architecture. The resultant predictions are highly accurate and tailored, potentially revolutionising the mobile app industry. Notably, both seq2seq modules in our model follow the "pre-train and fine-tune" approach using pre-trained weights from the HuggingFace Models repository\footnote{https://github.com/huggingface/transformers/tree/main/examples/pytorch/summarization}. This strategy, leveraging established linguistic patterns, significantly boosts prediction accuracy. Combining insights from both training stages, our model adeptly forecasts the next app for mobile users. Leveraging an array of contextual features and pre-trained LLMs, the Next App Prediction Module is a pivotal aspect of our approach. Our codes will be available for public access. Once ready, they can be found at the following repository.\footnote{https://github.com/cruiseresearchgroup/MAPLE.}

\color{black}

\section{EVALUATION}
This section aims to appraise our algorithm's efficacy utilizing authentic data derived from App usage. We present the methodology implemented to undertake the evaluation in Section \ref{sec:eval}. Subsequently, in Section \ref{sec:resultana}, we comprehensively examine the algorithm's performance. The evaluation is centred on two key objectives: a comparative analysis of the algorithm's performance relative to various baseline models and an exploration of the impact of each module incorporated in our proposed model.

\subsection{Evaluation setup}\label{sec:eval}
\subsubsection{Data Pre-Processing} The Tsinghua App Usage dataset used for evaluation contains over 2.3 million mobile app usage logs from 871 users in one week, from April 19, 2016, to April 26, 2016. The LSApp dataset used for evaluation contains around 600,000 mobile app usage logs from 293 users, with an average duration of 15 days per user. We separated the sessions of the app usage logs using a 5-minute threshold, and sessions with more than five thousand records were removed as noise. \hlnewtwo{We removed sessions with over five thousand records, considered as noise due to their repetitive pattern, likely resulting from automated processes or system glitches, not user behaviour.} Users with less than ten records were also removed at this stage. In our evaluation, we will consider the last fifteen applications used by a user as their historical app usage sequence. This sequence will provide the necessary context for our model to make predictions based on the user's app usage patterns. We will evaluate our proposed model with two different settings (Fig. \ref{fig:datasplit}), as shown below:
\begin{itemize}
    \item \hlnewtwo{Standard Setting: Following common practice in app prediction studies, we divided each user's data chronologically: 70\% for training, 10\% for validation, and 20\% for testing. This approach trained, fine-tuned, and evaluated our model's predictive precision}
    \item \hlnewtwo{Cold Start Setting: We tested our model's cold start proficiency by splitting users into two distinct groups: 90\% for training and 10\% of unseen users for testing. This method evaluated the model's performance with new users, addressing the cold start problem. The process was reiterated across all users to validate its effectiveness consistently.}
\end{itemize}

\hlnewtwo{Upon partitioning the dataset according to the specified settings, we subsequently employ a sliding window procedure to generate the samples.}

\begin{figure}[!h]

  \includegraphics[width=0.65\linewidth]{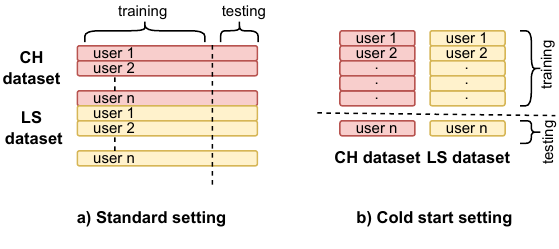}
  \caption{Comparison of Data Division Strategies for Standard and Cold Start Settings}
  \label{fig:datasplit}
\end{figure}

\hlcolor

\subsubsection{Performance Metric}
We assess the efficacy of our predictions through the utilization of the Accuracy@k and mean reciprocal rank (MRR) metrics \cite{kang2022app, khaokaew2021cosem}. Accuracy@k (A@K), computed as the mean of hits@k across all test predictions, yields a value of 1 if the actual app is among the top k predictions and 0 otherwise.

\begin{equation}
    A@k = \frac{\sum_{i=1}^{|D_{test}|}\alpha}{|D_{test}|},  \alpha| = \left\{ \begin{array}{cl}
1 & : \ y_{real} \in Y_{predict}\\
0 & : \ y_{real} \notin  Y_{predict}
\end{array} \right.
\end{equation}
where $|D_{test}|$ represents the number of test cases.

Meanwhile, MRR determines the reciprocal of the position at which the first relevant document was retrieved. The value of MRR is defined as follows, given the total number of items in the testing set (N):
\begin{equation}
MRR = \frac{\left( \sum_{i=1}^{|D_{test}|} \frac{1}{rank_{i}}\right)}{|D_{test}|}
\end{equation}
Where '$rank_{i}$' represents the position of the correct result in the 'i-th' prediction. Higher MRR values imply more accurate predictions. In our seq2seq model, we adapt these metrics to evaluate top k predictions by generating sentences until the top k predictions are collected.

\subsubsection{Baselines}
We compare our algorithm design with the following baselines to illustrate our proposed model's advantages.

\begin{itemize}
    \item[] \textbf{Vanilla model} - These are basic statistical baselines commonly employed in numerous \cite{tian2021and,zhou2020graph} to serve as a starting point for performance comparisons.
    \item MFU (Most Frequently Used): refers to the app that was the most frequently used by each user. 
    \item MRU (Most Recently Used): refers to the app that was the most recently used
by each user
    \item[] \textbf{App prediction model} - These neural network baselines are tailored for app usage prediction, leveraging techniques and architectures to capture app usage patterns and context, improving prediction accuracy and personalization.

\item AppUsage2Vec \cite{appusage2vec}: Deep learning model considering apps' combined influence, user characteristics, and usage context.
\item DeepApp \cite{xia2020deepapp}: Multitask learning model mapping time, location, and application to embeddings.
\item NeuSA \cite{aliannejadi2021context}: LSTM-based model considering app sequences and temporal user behaviour.
\item DeepPattern \cite{suleiman2021deeppatterns}: Spatiotemporal and context-aware model adapted from DeepApp.
\item CoSEM \cite{khaokaew2021cosem}: Model combining app seq. with semantic info.

    \item SA-GCN \cite{SA-GCN}: This model uses a Graph Convolutional Network \cite{kipf2016semi} to generate dense embeddings of apps, location, and time from an App usage graph, employing a "meta-path-based objective" for enriched representations.\footnote{Note that we use the same record count as other baselines for user vectors and employ location data from the public version, differing from the version in their paper.}
    \item[] \textbf{Time series forecasting models}\footnote{https://github.com/thuml/Time-Series-Library/}   - We also incorporate time series forecasting models in our experiments, as our historical app usage sequences resemble time series data. These models excel at detecting patterns in sequential data, serving as a pertinent baseline for our study's comparison.

    \item Transformer \cite{vaswani2017attention}: Encodes and decodes sequences with a self-attention mechanism.
\item FEDformer\cite{zhou2022fedformer}: Combines Fourier analysis with Transformer for long-term forecasting.
\item Reformer \cite{kitaev2020reformer}: Efficient Transformer model variant employing locality-sensitive hashing and reversible residual layers.
\item DLinear \cite{zeng2022transformersdliner}: Time series forecasting method combining a decomposition scheme with a linear model.
\item TimesNet \cite{wu2022timesnet}: Neural network architecture transforming 1D time series into 2D tensors.
    \item[] \textbf{Proposed model} 
\item MAPLE-ED: This is our proposed model that trains one model for each dataset. 
\item MAPLE: This is our proposed model, which considers all components and is trained with the samples from both datasets.
\end{itemize}

Note that the experiment with DeepApp and DeepPattern are only reported for the Tsinghua App Usage dataset because the LSapp dataset does not contain the location information, an essential feature in the DeepApp and DeepPattern frameworks.

\subsection{Result Analysis}\label{sec:resultana}
\subsubsection{Standard Setting} In this section, we compare the performance of our proposed model with the baseline mentioned in the previous section. We assess various models on the Tsinghua App Usage and LSapp datasets, as depicted in table \ref{tab:result data}. This comparison sheds light on their efficacy in app prediction tasks. The table delineates the performance of different models in predicting mobile app usage under standard conditions. MRR@K (M) and Accuracy@K (H) are performance metrics. Bold figures indicate the top performance, whereas underlined figures denote the second best. The evaluation results demonstrate that our proposed models, MAPLE and MAPLE-ED, consistently surpass other models across both Tsinghua App Usage and LSapp datasets in most metrics. This success underscores the robustness of our approach in addressing the app prediction challenge.

\color{black}

\begin{table}[!h]
  \small
  \caption{Performance Comparison in Standard Setting (*M=MRR@K, A=Accuracy@K) \\ The bold values represent the best performance, while the underlined values indicate the second-best performance. }
  \label{tab:result data}
  \begin{tabular}{l|ccccc|ccccc}
    \toprule
    Dataset & \multicolumn{5}{c|}{Tsinghua App Usage}&   \multicolumn{5}{c} {LSapp}\\
    \midrule
    Model/Metric & A@1 &A@3 & A@5&M@3&M@5 &A@1 &A@3 & A@5&M@3&M@5\\
    \midrule
   
MFU &0.1972 & 0.4288 & 0.5384 &  0.2991 & 0.3241 & 0.2952 & 0.6258 & 0.7942 & 0.4378 & 0.4765 \\
MRU& 0.000 & 0.5538 & 0.6536 &  0.2585 & 0.2817 &0.0276 & 0.7850 & 0.8306 &0.3974 & 0.4079 \\

\midrule

Appusage2Vec&0.2909 & 0.4822 & 0.5781&0.3739 & 0.3958 &  0.6057 & 0.7858 & 0.8618 & 0.6848 & 0.7022\\
NeuSA&0.4640 & 0.6562 & 0.7286 & 0.5492 & 0.5658&0.6832 & 0.8253 & 0.8830&0.7461 & 0.7593 \\

SA-GCN&0.0613&0.1882&0.2521&0.1183&0.1331&-&-&-&-&-\\
DeepApp&0.2862 & 0.5931 & 0.7075&0.4210 & 0.4473 &-&-&-&-&-\\
DeepPattern&0.2848 & 0.5884 & 0.7016 & 0.4185 & 0.4444 &-&-&-&-&-\\
CoSEM&0.4163 & 0.6682 & 0.7499& 0.5282 & 0.5469&0.4990 & 0.7466 & 0.8149& 0.6083 & 0.6242\\

\midrule
TimesNet&0.0208 & 0.048 & 0.0614 & 0.0327 & 0.0358&0.4805 & 0.628 & 0.6897 & 0.5459 & 0.5600\\
Transformer&0.0262 & 0.0534 & 0.0661& 0.0383 & 0.0412 &0.4978 & 0.653 & 0.7141 & 0.5659 & 0.5800\\
FEDformer&0.0159 & 0.042 & 0.0553& 0.0272 & 0.0303 &0.4946 & 0.6374 & 0.6915& 0.5585 & 0.5708\\
DLinear&0.0072 & 0.037 & 0.0607& 0.0202 & 0.0256 & 0.1611 & 0.3978 & 0.479& 0.2637 & 0.2824\\
Reformer&0.0228 & 0.0503 & 0.0645& 0.0346 & 0.0378 & 0.4920 & 0.6505 & 0.7074& 0.5620 & 0.575\\

\midrule

\midrule
\textbf{MAPLE-ED}& \underline{0.5173} & \underline{0.7349} & \underline{0.8070}  & \underline{0.6142} & \underline{0.6308} &
\textbf{0.7171} & \textbf{0.8670} & \textbf{0.9166} & \textbf{0.7836} & \textbf{0.7950} \\
\textbf{MAPLE}&\textbf{0.5191}& \textbf{0.7385} & \textbf{0.8115} & \textbf{0.6169} & \textbf{0.6338 }& \underline{0.7157}& \underline{0.8649} & \underline{0.9150}&  \underline{0.7821} & \underline{0.7936  }\\

  \bottomrule
\end{tabular}

\end{table}

Compared to basic benchmarks like MFU and MRU, specialized app prediction models such as AppUsage2Vec, DeepApp, NeuSA, DeepPattern, CoSEM, and SA-GCN perform better but don't reach the levels of our proposed models. NeuSA, with its LSTM structure, particularly excels at capturing user behaviour. However, time series models like Transformer, FEDformer, Reformer, DLinear, and TimesNet, despite being adapted for app prediction, fall short due to their primary focus on time series forecasting, not app context specifics. This distinction in focus, combined with the larger scale of the Tsinghua App Usage dataset, affects their performance, especially when compared to models designed specifically for app prediction tasks.

Our MAPLE and MAPLE-ED models excel in app prediction due to their use of a pre-trained LLM and cross-dataset user similarity. However, MAPLE's slightly weaker performance on the LSapp dataset may stem from its size discrepancy with the Tsinghua App Usage dataset and its incorporation of location data unavailable in LSapp. In essence, while MAPLE-ED stands out in the app prediction, emphasizing the power of LLMs and multi-dataset synergy, model selection should be attuned to specific use cases and dataset nuances. We'll next probe our model's mettle against the user cold-start challenge, laying the groundwork for deeper analysis in upcoming stages.

\subsubsection{User Cold Start Setting} In this section, we evaluate the performance of the models in the user cold start problem. It is important to note that we only selected baselines that do not require user information in their models, as we could not access data from the new users. Table \ref{tab:cold result data} presents the results for the Tsinghua App Usage and LSapp datasets. It is evident that our proposed MAPLE model outperforms all other models in terms of Accuracy and Mean Reciprocal Rank metrics for both datasets, emphasizing its effectiveness in handling the cold start problem.

\begin{table}[!h]
  \small
  \caption{Performance comparison Cold start setting (*M=MRR@K, A=Accuracy@K )}
  \label{tab:cold result data}
  \begin{tabular}{l|ccccc|ccccc}
    \toprule
    Dataset & \multicolumn{5}{c|}{Tsinghua App Usage}&   \multicolumn{5}{c} {LSapp}\\
    \midrule
    Model/Metric & A@1 &A@3 & A@5&M@3&M@5 &A@1 &A@3 & A@5&M@3&M@5\\
    \midrule
        MFU& 0.1853 & 0.3906 & 0.4943 & 0.2752 & 0.2989  & 0.6098 & 0.7716 & 0.7377&0.4295 & 0.4667 \\
    MRU&0.0000 & 0.6406 & 0.7226 &  0.3042 & 0.3234 & 0.0155 & 0.8625 & 0.8868 & 0.4341 & 0.4397  \\

    \midrule
    TimesNet&0.0144 & 0.0433 & 0.0647 &0.0277 & 0.0323 &0.0022 & 0.0114 & 0.0246&0.0059 & 0.0089\\
Transformer&0.0180 & 0.0461 & 0.0606 & 0.0308 & 0.0337 & 0.0028 & 0.0174 & 0.0702 & 0.0085 & 0.0201\\
FEDformer&0.0100 & 0.0411 & 0.061 & 0.0231 & 0.0282 & 0.0024 & 0.0075 & 0.0231 & 0.0044 & 0.0081\\

Reformer&0.0224 & 0.0506 & 0.057&0.0349 & 0.0384 &0.0034 & 0.0132 & 0.0330 &  0.0064 & 0.0119 \\
DLinear&0.0070 & 0.035 & 0.0586&0.0199 & 0.0245 &  0.1600 & 0.3968 & 0.4770 & 0.2617 & 0.2814\\
    \midrule

    CoSEM&0.3111 & 0.5597 & 0.6525 &  0.4204 & 0.4416 & 0.4523 & 0.7243 & 0.8104 & 0.5718 & 0.5918  \\ 
        NeuSA  & 0.4433 & 0.6169 & 0.6812& 0.5206 & 0.5353 &  0.6874 & 0.777 & 0.8135& 0.7272 & 0.7355  \\
         \midrule
          \midrule

    MAPLE w/o ins. app & 0.5227 & 0.7405 & 0.8060  & 0.6201 & 0.6352 & 0.7645 & 0.8718 & 0.9061  & 0.8128 & 0.8207 \\
     MAPLE &\textbf{0.5228} & \textbf{0.7417} & \textbf{0.8128} & \textbf{0.6206} & \textbf{0.6369} &  \textbf{0.7644 }& \textbf{0.8848 }& \textbf{0.9247} & \textbf{0.8181} & \textbf{0.8272 }\\

    \midrule

  \bottomrule
\end{tabular}

\end{table}

\hlnewtwo{
The exceptional performance of our MAPLE model is attributed to its integration of users' installed app datasets. MAPLE analyzes app installation patterns on mobile devices to deduce user behaviors and preferences. This feature is particularly advantageous for new users lacking prior usage history. Our hypothesis was confirmed through a performance comparison between our model with installed app data (MAPLE) and without it (MAPLE w/o ins. app). Results indicated that although the modified model was effective, the complete version's employment of installed app data markedly improved predictive accuracy.}

\hlnewtwo{
Furthermore, MAPLE's deployment of a pre-trained LLM enables it to adapt swiftly and more accurately to new datasets, demonstrating its robustness in scenarios where typical LLMs may struggle. This is in contrast to time series forecasting models that depend on a consistent data distribution across training and test sets—a condition that is rarely met with new user data. While models like NeuSA and CoSEM show potential by leveraging historical app usage patterns, they fall short of MAPLE's performance, which gains from a comprehensive understanding of user behaviour gleaned from historical usage and installed apps.}

In summary, the results underscore the robustness and efficacy of our MAPLE model in tackling the user cold start problem, significantly surpassing other models, including established app prediction and time series forecasting approaches. These findings confirm that the MAPLE model's nuanced ability to leverage data from installed apps, identify similar user profiles, and utilise a pre-trained LLM is well-equipped to manage the complexities associated with new user predictions effectively.

\hlcolor

\subsubsection{Analysis of Model Parameters}

\begin{figure}[h!]

     \begin{subfigure}[b]{0.45\textwidth}
         \centering
           \includegraphics[width=\textwidth]{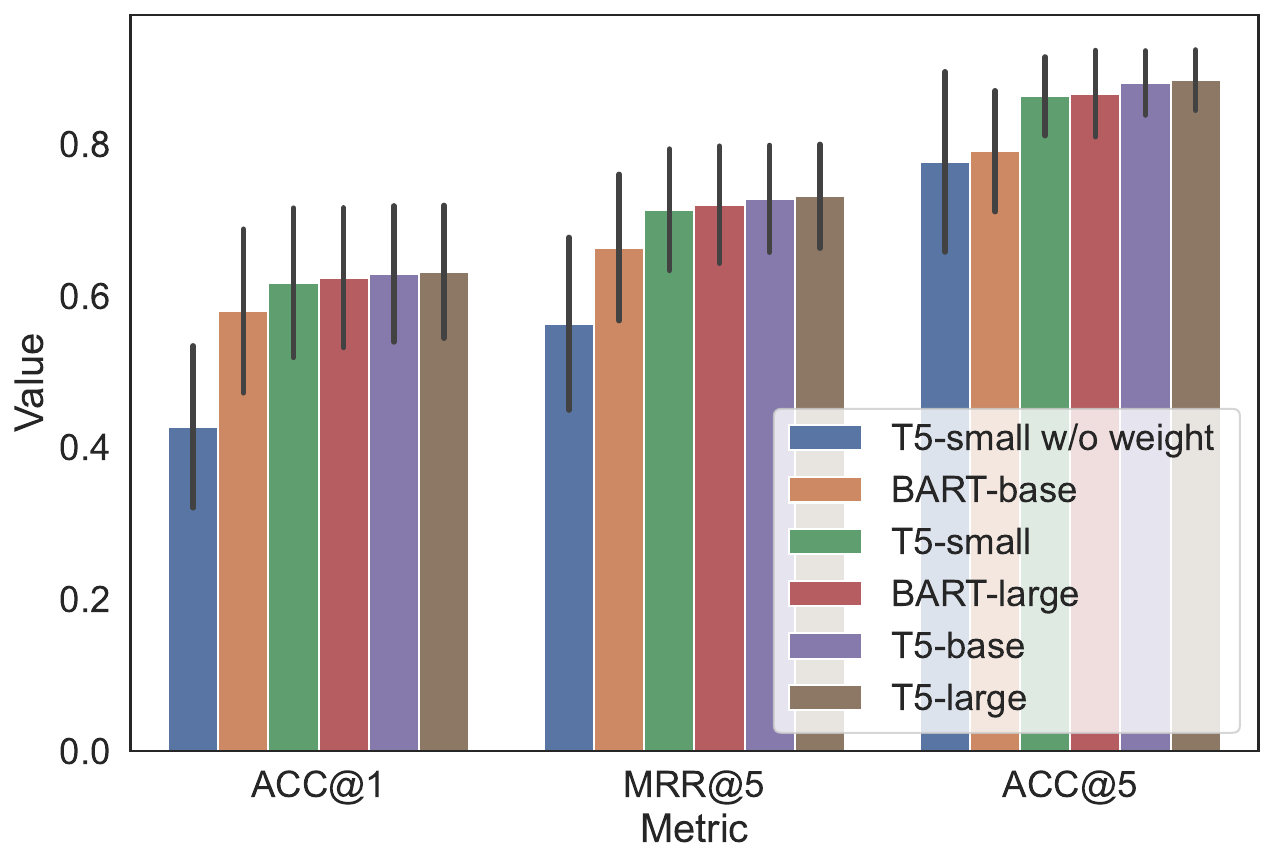}
         \caption{MAPLE with different pre-trained model}
         \label{fig:diff pretrain}
     \end{subfigure}
          \begin{subfigure}[b]{0.45\textwidth}
         \centering
           \includegraphics[width=\textwidth]{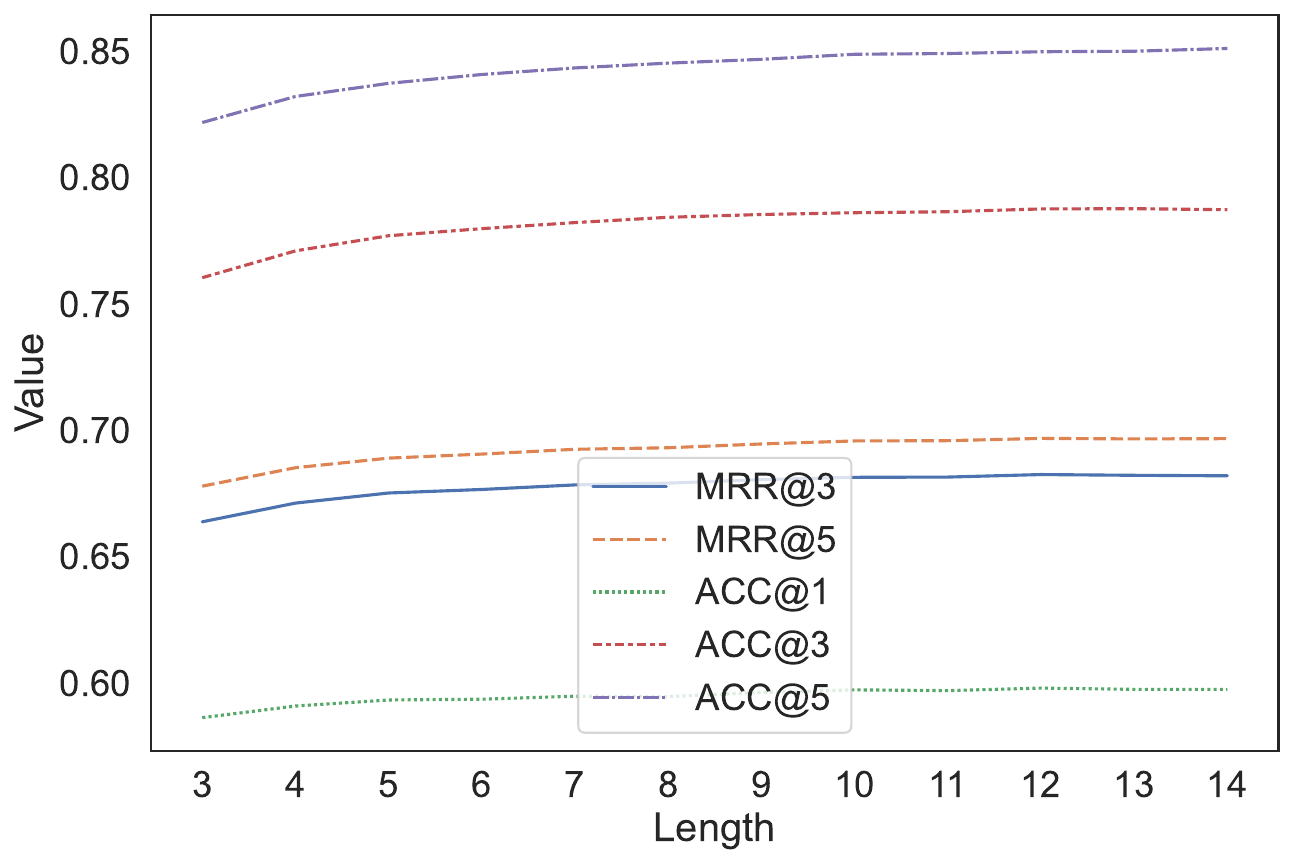}
         \caption{MAPLE with varying historical app usage length}
         \label{fig:diff_lenght}
     \end{subfigure}

  \caption{Performance of MAPLE with different parameters}
  \label{fig:app len}
\end{figure}
In this section, we analyze the performance of the proposed model on different parameters. Specifically, we investigate the effect of using different pre-trained language models and varying the length of the historical app usage sequences.

Our study assessed the efficacy of various renowned pre-trained language models, including T5 and BART, by integrating them into our MAPLE framework and scrutinizing performance through pertinent metrics. As depicted in Figure \ref{fig:diff pretrain}, performance diverges across the models. Notably, T5-large leads in Accuracy@1, Accuracy@5, and MRR@5 for both datasets.

Further examination reveals a clear positive correlation between the size of pre-trained models and the performance of MAPLE. Both T5 and BART demonstrate improved results as we scale up from smaller to larger models, indicating the larger models' superior ability to interpret complex app usage patterns. Additionally, the efficacy of pre-training strategies is evident. T5, with its denoising autoencoder objective and multi-task training approach, consistently outperforms BART’s sequence-to-sequence framework. This suggests that a multi-tasking pre-training regimen might be more suitable for app usage prediction. The notable difference in performance between the T5-small models, with and without pre-trained weights, also emphasizes the critical role of knowledge transfer via pre-trained models in this domain. These findings solidly affirm the vital role of advanced pre-trained language models in boosting predictive accuracy for app usage within the MAPLE framework.

In evaluating our proposed model alongside various pre-trained language models, we also examined the impact of varying historical app usage sequence lengths, from 3 to 14. The findings indicate that longer historical sequences correspond to improved model performance. Notably, metrics like MRR@5 and ACC@5 showed enhancement with extended sequences, illustrating the model's increased proficiency in detecting app usage patterns. However, this trend levels off after approximately 11-12 sequences, hinting at an optimal sequence length that balances comprehensive historical data with computational efficiency.

In conclusion, our analysis of the MAPLE model, focusing on the choice of pre-trained model and the length of historical sequences, reveals that employing models such as T5-large and fine-tuning the sequence length can significantly refine app usage predictions. This insight is instrumental in developing advanced prediction models that more accurately capture app usage patterns and enhance the precision of forecasts.

\begin{table*}[!h]
  
  \caption{Ablation study}
  \label{tab:abstudy data}
  \begin{tabular}{l|ccccc|ccccc}
    \toprule

   \textbf{ Dataset} & \multicolumn{5}{c|}{\textbf{Tsinghua App Usage}}&   \multicolumn{5}{c} {\textbf{LSapp}}\\
    \midrule
    \textbf{Model/Metric} & A@1 &A@3 & A@5&M@3&M@5 &A@1 &A@3 & A@5&M@3&M@5\\

    \midrule

    w/o $1^{st}$ stage training & 0.3226 & 0.5598 & 0.6529  & 0.4275 & 0.4488 &  0.5326 & 0.8099 & 0.8922 & 0.6561 & 0.6750  \\
    
    w/o App seq info & 0.3146 & 0.5393 & 0.6392  & 0.4133 & 0.4361 & 0.4853 & 0.7693 & 0.8549  & 0.611 & 0.6306   \\
    w/o installed App  & 0.5091 & 0.7221 & 0.7892  & 0.6042 & 0.6196 & 0.7146 & 0.8554 & 0.8996 & 0.7772 & 0.7874  \\
    w/o optional contexts& 0.5135 & 0.7314 & 0.8035 &  0.6105 & 0.6272 & 0.7145 & 0.8626 & 0.9126 & 0.7804 & 0.7920   \\
    \midrule
    \textbf{MAPLE} &\textbf{0.5191}& \textbf{0.7385} & \textbf{0.8115} & \textbf{0.6169} & \textbf{0.6338 }& \textbf{0.7157}& \textbf{0.8649} & \textbf{0.9150}&  \textbf{0.7821} & \textbf{0.7936  }\\

  \bottomrule
\end{tabular}

\end{table*}

\subsubsection{Ablation Study} In this section, we present an ablation study to precisely delineate the individual contribution of each component in the MAPLE model. We aim to understand each module's weightage in contributing to our model's overall effectiveness. We tested the following model configurations:

\begin{itemize}

    \item MAPLE w/o $1^{st}$ stage training is our model trained solely on the second stage without leveraging the app-type prediction training.
    \item MAPLE w/o App seq info is our proposed model without using historical app usage information.
    \item MAPLE w/o installed Apps is our proposed model, which does not include the set of installed sentences.
   \item MAPLE w/o optional context is a model that omits the optional context (location context - POI in our cases),
\end{itemize}

Table \ref{tab:abstudy data} presents a detailed overview of performance metrics for various configurations of the MAPLE model across two datasets: Tsinghua App Usage and LSapp. Each configuration corresponds to a version of the MAPLE model with a particular component excluded, aimed at isolating its individual impact.

In the "w/o $1^{st}$ stage training" configuration, the omission of the initial app-type prediction stage leads to a decline in performance. For the Tsinghua App Usage dataset, the model yields an MRR@1 of 0.3226 and Accuracy@5 of 0.6529, both lower than the complete MAPLE model. Similarly, for the LSapp dataset, the metrics are 0.5326 and 0.8922, respectively, demonstrating a performance decrease. When excluding the historical app usage data, denoted as "w/o App seq info," the model once again exhibits inferior performance, with an MRR@1 of 0.3146 and Accuracy@5 of 0.6392 for the Tsinghua App Usage dataset, highlighting the significance of this component.

\hlnewtwo{
The interaction of our model with Point of Interest (POI) data in the "w/o optional contexts" setup provides crucial insights. Omitting this context impacts the model's effectiveness. For example, in the Tsinghua App Usage dataset without POI data, MRR@1 and Accuracy@5 are 0.5135 and 0.8035, respectively. In contrast, with the full MAPLE model including all contexts, these metrics improve to 0.5191 and 0.8115. Notably, even in the LSapp dataset, which lacks explicit POI information, an improvement is seen. This suggests the model learns to identify app usage patterns indicative of POI-related behavior during training with Tsinghua App Usage data. When it encounters similar patterns in LSapp, it uses these associations to improve predictions. However, this approach has limitations. The model might sometimes misinterpret LSapp behaviors as influenced by POI, causing occasional inaccuracies, as seen in the MAPLE-SEP's performance.}

In conclusion, the MAPLE model's ability to integrate various contexts and maintain performance despite missing elements demonstrates its robustness. However, careful consideration is required to ensure optimal accuracy, particularly when interpreting patterns from diverse datasets.

\color{black}

\subsubsection{Limitation on resource requirements}

\begin{table}[!h]
  \caption{Memory usage analysis}
  \label{tab:memory}
  \begin{tabular}{l|ccccc}
    \toprule

    Model/Metric & \# Parameters  & Model size (b)  & \multicolumn{1}{>{\centering\arraybackslash}p{2.5cm}}{Inference time (s/samples)}& \multicolumn{1}{>{\centering\arraybackslash}p{1.9cm}}{H@1 Tsinghua App Usage}& \multicolumn{1}{>{\centering\arraybackslash}p{1.5cm}}{ H@1 LSApp}\\

    \midrule

    DeepPattern& $\approx11M$ & $\approx42 M$ & $0.1267e-2$&0.2848 &- \\
    DeepApp& $\approx10.8M$ & $\approx42 M$ & $0.0947e-2$&0.2862 &-\\
    Appusage2Vec& $\approx3.8 M$ & $\approx40 M$ &$0.0829e-2$& 0.2909 &0.6057\\
    CoSEM& $\approx8.8 M$ & $\approx35 M$& $0.0790e-2$ & 0.4163& 0.4990 \\ 
        NeuSA  & $\approx28M$ & $\approx167 M$&  $0.1197e-2$ & 0.4640&0.6832 \\ 
         \midrule

    \textbf{MAPLE (T5-small)} & $\approx60 M$ &  $\approx480M$ &$1.1270e-2$& 0.5191&0.7157 \\

  \bottomrule
\end{tabular}

\end{table}
\hlnewtwo{ The memory usage analysis of top-performing models in app usage prediction is captured in Table \ref{tab:memory}, highlighting not only the predictive accuracy as measured by Hit Rate at 1 (H@1) but also the computational resources each model demands, a critical factor for real-world application. Our proposed MAPLE model, leveraging a smaller variant of the T5, demonstrates exceptional predictive performance with the highest H@1 scores for both the Tsinghua App Usage and LSApp datasets. However, it is also the most resource-intensive in terms of the number of parameters, model size, and inference times.}

\hlnewtwo{ While MAPLE's considerable resource demands, evident in its inference time, may present challenges in settings with stringent memory or processing speed requirements, this is offset by the evolving dynamics of LLMs and advancements in hardware efficiency, which increasingly support the use of more robust models. The versatility of LLMs like T5, which are capable of handling multiple tasks, argues in favor of their adoption, as the overhead can be spread over a range of applications, improving the cost-effectiveness. The trend towards integrating advanced LLMs into mobile technology is anticipated to accelerate, driven by ongoing improvements in model optimization and hardware performance. This integration is expected to enhance not just app usage prediction but also to equip devices with a broad array of intelligent functions, utilizing the comprehensive potential of LLMs. Employing multi-functional LLMs can rationalize the initial resource expenditure by delivering a more adaptable and potent system on mobile platforms.}

\hlnewtwo{In conclusion, the MAPLE model's exceptional performance, despite its resource intensity, underscores the benefits of LLMs in app usage prediction. Our ongoing research will focus on improving computational efficiency and examining the broader applications of LLMs, reinforcing our commitment to developing scalable and multifunctional AI systems for mobile devices.}

\section{DISCUSSION and IMPLICATION}

\hlcolor

Our model outshines traditional mobile app prediction methods using LLMs and contextual data to offer more precise and tailored predictions. The ATP and NAP Modules, using personal app histories and installed apps, along with context, lead to accurate predictions of a user's next app. This approach allows for a richer understanding of user patterns, enabling finer predictions. By harnessing LLMs, our model expertly deciphers correlations between apps and contextual factors—like location and time—boosting its predictive success. Our Two-Stage LLM Training Module fine-tunes these insights to mirror individual behaviours, enhancing user experiences for app developers and suggesting new horizons for app recommendations. Our approach adeptly addresses the user cold start issue and capitalizes on app data to elevate prediction personalization, showcasing the potential for wider applications and setting the stage for future innovation.

One important implication of our work is that it provides a framework for developing more effective and personalised mobile app recommendation systems. By incorporating contextual information and personal app history data, our proposed model offers a more comprehensive approach to app recommendation that considers the user's current context and the app usage patterns from multiple large datasets. This can lead to more engaging and enjoyable mobile user experiences, ultimately increasing user satisfaction and retention.

Another implication of our work is that it demonstrates the power of LLM-based models for handling complex, heterogeneous data. Our model combines contextual information with personal app history data, which can be highly heterogeneous and difficult to model using traditional machine learning algorithms. By leveraging the power of LLM, we can model the complex relationships between the user's app usage patterns and contextual data, resulting in more accurate and personalised app recommendations.

\hlnewtwo{One potential extension of our proposed model is to modify the next app prompt to include more types of target variables, such as the expected time of usage for the next app ("3" seconds from now, this user will use the "Facebook" app). Our model could offer users more comprehensive and personalised recommendations by incorporating additional variables into the next app prompt. This has important implications for many applications beyond mobile next-app prediction, including time management and productivity. For example, our model could be used to recommend apps that are most likely to be used during certain times of the day based on the user's historical app usage patterns. Furthermore, our proposed model has the potential to identify new apps that are highly relevant to the user's interests and usage patterns by leveraging both personal app history data and contextual information. For example, if a user frequently uses apps related to health and fitness, our model could recommend a new app in that category that matches the user's preferences. This has important implications for app developers, who could use our model to introduce their apps to users in a more personalised and targeted way.}

Overall, our proposed model offers a promising approach to mobile app usage prediction that combines the strengths of LLM-based models with contextual information, personal app history data, and the set of installed applications. By incorporating these data types, our model offers a more accurate and personalised approach to app recommendation, improving the overall user experience for mobile users. We believe that our work has important implications for developing more effective and personalised recommendation systems across a wide range of domains.

\color{black}

\section{RELATED WORKS}

\hlcolor

\subsection{App usage prediction}
Over the past decade, the surge in mobile users has elevated interest in app usage prediction. \citet{Huang_ubicome2012} utilised Bayesian methods, incorporating features such as location and mobile state, and achieved a 69\% accuracy on the Nokia MDC dataset \cite{laurila2012mobile}. Meanwhile, \citet{Shin_Ubicomp2012} introduced a naive Bayes-based model, employing various smartphone contextual data, which realized 78\% accuracy. Emphasizing efficient feature usage, \citet{Liao_2013_ICDM} employed feature selection with a kNN classifier, crafting an app usage graph to attain 80\% recall at top-5 predictions. \cite{Xu_appbag_ISWC2013} explored the influence of social contexts by leveraging community behaviour similarities for prediction improvement. Conversely, \citeauthor{parate_ubicom_appm_2013}\cite{parate_ubicom_appm_2013} demonstrated an 81\% accuracy by focusing solely on app usage history, asserting that privacy-sensitive features are non-essential. Similarly, \citeauthor{Natarajan_2013} \cite{Natarajan_2013} underscored the relevance of the last app used in a session, using a cluster-level Markov model and the iConRank algorithm to delve into app transition behaviours.

\citeauthor{Baeza-Yates.yahoo.2015.WSDM}\cite{Baeza-Yates.yahoo.2015.WSDM} used the idea from the Word2vec \cite{word2vec} approach for extracting the app session feature from the app usage record. This app session features will be used as input in a parallelized Tree Augmented Naive Bayes model (PTAN) with other contextual features. The contextual features and the app session features are related to each other in the app usage prediction, however, the PTAN model assumes these features to be independent. \citeauthor{appusage2vec} \cite{appusage2vec} proposed the Appusage2vec in 2019. Their work introduced an app-attention mechanism to measure the contribution of each app to the target app and also proposed the Dual Deep Neural Network model that can learn the user vector from the app usage record. Their model can achieve the 84.47 \% of Recall@5 on the large app usage dataset.

In addition, \citeauthor{Aliannejadi.CIKM201810.1145/3269206.3271679} \cite{Aliannejadi.CIKM201810.1145/3269206.3271679} introduced a framework for predicting target apps based on search intent and contextual features, leading to the creation of a universal mobile search system. In contrast, \citeauthor{Cap} \cite{Cap} presented the CAP model to forecast app usage across various locations and times by forming a relationship vector among app, location, and time. \citeauthor{Wang2019_ubicom_appusage}\cite{Wang2019_ubicom_appusage} addressed data sparsity in large datasets by creating a spatio-temporal app usage model, focusing on predicting app categories rather than specific apps. To enhance app prediction and address data scarcity in new locations, \citeauthor{Fan_2019_transfer_learning}\cite{Fan_2019_transfer_learning} devised a transfer model using data from both app usage and check-in datasets. Finally, \citet{kang2022app} developed a model for predicting app usage during specific contexts like commuting, highlighting the importance of contextual information for accurate predictions when users are on the move.

These previous studies have demonstrated the effectiveness of using contextual information to enhance the modelling performance of the app usage prediction model. The proposed models can solve the app usage prediction problem using a specific type of contextual feature and provide high performance. In some situations, however, these types of contextual features may not be available. Therefore, we develop our model that can be applied to different types of contextual information and address the contextual data heterogeneity problem.  

\subsection{Large language model on other domains}

Large language models (LLMs) have brought advancements across various fields, including natural language processing (NLP), healthcare, recommender systems, coding assistance, multimodal learning, and human mobility forecasting, demonstrating their flexibility and potential. In NLP, LLMs like BERT \cite{devlin2018bert}, GPT \cite{radford2019languagegpt}, and RoBERTa \cite{liu2019roberta} have set benchmarks in tasks such as sentiment analysis, translation, and summarization. They have enabled fine-tuning for specific tasks, such as enhancing performances in text classification \cite{gonzalez2020comparingtextclassification}.

The healthcare sector has also embraced LLMs, utilizing them in medical information extraction and patient triage. For example, BioBERT \cite{lee2020biobert}, a domain-specific LLM based on BERT, has been pre-trained on a large-scale biomedical corpus and employed for tasks like named entity recognition, relation extraction, and question-answering within the biomedical literature. Moreover, the recommender systems domain has also integrated LLMs to capture complex patterns in user behaviour and contextual information. For instance, BERT4Rec \cite{sun2019bert4rec} is adept at capturing user behaviour in a session-based setting, surpassing traditional methods. LLMs have also been used to model user-item interactions and incorporate temporal dynamics, improving recommendation personalization. In the realm of code generation, LLMs such as OpenAI's Codex \cite{chen2021evaluatingcodex,finnie2022robotscodex} and CodeBERT \cite{feng2020codebert} have been valuable in generating code snippets, suggesting API usage, performing code completion tasks, and assisting in debugging.

LLMs have been instrumental in multimodal learning, which involves fusing information from different data modalities, such as text, images, and videos. Models like ViLBERT \cite{lu2019vilbert} have performed tasks like visual question answering and visual commonsense reasoning, showing LLMs' ability to reason across diverse data types. Temporal sequential pattern mining is another domain where LLMs have shown potential, as in the AuxMobLCast \cite{xue2022leveragingmobicast} pipeline that predicts Place-of-Interest customer flows, demonstrating their capability to model human behaviour.

Despite LLMs’ extensive applications, their use for app usage prediction remains underexplored. Given LLMs’ proven aptitude in domains like mobility forecasting, we assert that their ability to discern complex user behaviours and contexts can enhance app usage predictions. Consequently, we propose an innovative LLM-based method for app usage prediction, expanding LLM research and showcasing their potential to tackle challenges inherent to this task.

\section{Conclusion}

In conclusion, this research paper presents a novel approach for predicting app usage behaviour by leveraging language foundation models (LLMs). Our model, MAPLE, outshines existing benchmarks in standard and cold start scenarios, encompassing both time series forecasting models and various other app prediction models. The integration of LLMs enables our model to adeptly adjust to different datasets through pre-trained models, taking advantage of the data available in both datasets and the array of apps installed on users’ phones. Incorporating installed apps empowers our model to address the user cold start issue, as it identifies similar users through their app installation patterns, potentially reflective of akin habits and lifestyles. This key information is then leveraged to anticipate app usage behaviour for new users. 

One of the main reasons for the success of our proposed model is the ability of LLMs to capture complex patterns and relationships within data. In the realm of app usage prediction, this translates to a deeper understanding of user behaviour and context, resulting in more accurate forecasts. The related works discussed in this paper also underscore the versatility and applicability of LLMs across diverse domains, such as human mobility forecasting and sentiment analysis. These explorations not only corroborate the efficacy of LLMs but also pave the way for their potential application in a broader range of predictive modelling scenarios, extending their relevance beyond the scope of our current study.

\hlnewtwo{Future research opportunities abound in enhancing the predictive prowess of our model. Investigating various LLM architectures alongside novel fine-tuning methodologies holds promise for boosting performance. Delving into the integration of richer contextual datasets, including user demographics and broader environmental factors, could unveil deeper influences on app utilization patterns. Additionally, the prospect of developing specialized, lightweight LLMs presents an enticing pathway to reduce computational demands while maintaining accuracy. The adaptability of our model also paves the way for its application across diverse domains with sequential data, potentially expanding its utility and significance in user behaviour prediction.}

In summary, this research paper has successfully demonstrated the potential of LLMs in app usage prediction and has opened up new opportunities for future research in this area. With its unique inclusion of the set of installed apps, the proposed MAPLE model has shown promising results in both regular and cold start settings, highlighting the value of leveraging LLMs in understanding and modelling human behaviour across various domains.

\section*{ACKNOWLEDGMENTS}
This research is supported by the Royal Thai Government, the UNSW RTP scholarship, and the ARC Centre of Excellence for Automated Decision-Making and Society (CE200100005). Additionally, this research was undertaken with the assistance of resources and services from the National Computational Infrastructure (NCI), which is supported by the Australian Government.

\color{black}


\bibliographystyle{ACM-Reference-Format}
\bibliography{ref}




\end{document}